\documentclass[runningheads]{llncs}

\usepackage{eccv}

\usepackage{eccvabbrv}

\usepackage{graphicx}
\usepackage{booktabs}

\usepackage[accsupp]{axessibility}  

\usepackage{hyperref}
\usepackage{url}        
\usepackage{orcidlink}

\usepackage{amsfonts}       
\usepackage{nicefrac}       
\usepackage{microtype}      
\usepackage{xcolor}         
\usepackage{wrapfig}
\usepackage{tikz}
\usepackage{subcaption}
\usepackage{multirow}
\usepackage{adjustbox}
\usepackage{pgfplots}
\pgfplotsset{compat=1.18} %
\usepackage{amsmath}

\usepackage{amsthm}
\usepackage{algorithm}
\usepackage{algorithmic}
\usepackage{amssymb}
\usepackage{minitoc}
\usepackage[toc,page,header]{appendix}
\usepackage{titletoc} 
\usepackage{dsfont}

\usepackage{tabularx}
\usepgfplotslibrary{groupplots}
\theoremstyle{plain}

\newtheorem{assumption}{Assumption}
\theoremstyle{remark}

\begin{document}

\title{GeoDetect: Geometric Adversarial Detection for VLPs} 

\titlerunning{GeoDetect}

\author{Afsaneh Hasanebrahimi\inst{1}\orcidlink{0009-0002-6088-1543}\and
Hanxun Huang\inst{1}\orcidlink{0000-0002-2793-6680}\and
Christopher Leckie\inst{1}\orcidlink{0000-0002-4388-0517}\and
James Bailey\inst{2}\orcidlink{0000-0002-3769-3811}\and
Sarah Erfani\inst{1}\orcidlink{0000-0003-0885-0643}}

\authorrunning{A.~Hasanebrahimi et al.}

\institute{
The University of Melbourne, Melbourne, Australia\\
\email{\{a.hasanebrahimi, curtis.huang1, caleckie, sarah.erfani\}@unimelb.edu.au}
\and
Monash University, Melbourne, Australia\\
\email{james.a.bailey@monash.edu}
}

\maketitle

\begin{abstract}
Vision-language pre-trained models (VLPs) are widely used in real-world applications. However, they remain vulnerable to adversarial attacks. Although adversarial detection methods have demonstrated success in single-modality settings (either vision or language), their effectiveness and reliability in multimodal models such as VLPs remain largely unexplored. In this work, we study the geometry of VLP embedding spaces and observe structured anisotropy that differs from unimodal vision models. Our theoretical analysis shows that under this anisotropic structure, adversarial attacks increase the expected geometric separation between clean and adversarial examples (AEs).
Specifically, we demonstrate that AEs consistently exhibit greater expected distances to randomly sampled points than their clean counterparts, indicating that AEs tend to push representations out of manifold regions. 
Building on these insights, we propose GeoDetect, which leverages these off-manifold deviations via geometric scores to identify AEs. Through comprehensive evaluations, we show that our approach reliably detects AEs across diverse VLP architectures and threat settings, covering unimodal and multimodal attacks as well as adaptive attacks, thereby providing a robust and practical approach to improving the safety and reliability of these models.
\end{abstract}

\section{Introduction}

Vision-language pre-trained models (VLPs) enable the understanding of both visual and textual data by learning joint representations of multimodal inputs. This capability makes them highly effective for tasks requiring a deep understanding of 
images and text. VLPs have achieved state-of-the-art results across various multimodal tasks \cite{yin2023survey, xu2023multimodal, gandhi2023multimodal}, including image-text retrieval \cite{chen2020imram}, visual question answering 
\cite{lu2019vilbert}, and zero-shot classification \cite{radford2021clip}. Despite their remarkable success, VLPs remain vulnerable to adversarial examples (AEs) \cite{zhang2022towards, schlarmann2023adversarial}, raising concerns about their robustness in real-world, safety-critical applications. 

Recent research has explored adversarial training as a strategy to enhance the zero-shot robustness of VLPs \cite{mao2022understanding, wang2024pre, schlarmann2024robust}. However, adversarial training is computationally expensive \cite{madry2017towards, wang2019improving} and often involves a trade-off between model performance and robustness \cite{zhang2019theoretically, tsipras2018robustness}. Detecting AEs presents a more flexible alternative by allowing the model to identify and reject potentially harmful queries, rather than attempting to provide reliable outputs for all inputs. 

While existing work has been proposed for detecting AEs in unimodal models \cite{feinman2017detecting, lee2018simple, ma2018characterizing, sotgiu2020deep, kherchouche2020detection, aldahdooh2023revisiting}, it remains uncertain whether detection signals used in unimodal detectors can transfer to VLPs, since multimodal alignment and fusion can reshape the geometry of clean neighborhoods in embedding space. 
In contrast to standard unimodal classification models trained with cross-entropy loss to predict discrete labels, VLPs are optimized to align image and text representations within a shared embedding space using contrastive learning objectives \cite{radford2021clip}. Recent findings by \cite{schlarmann2024robust} demonstrate that CLIP embeddings experience significant distortion under adversarial attack, as evidenced by substantial shifts in the embedding space.  
A recent study by \cite{zhang2024pip} proposed Prompt-based Irrelevant Probing (PIP), a task-specific detection method for visual question answering (VQA) that analyzes attention responses to irrelevant probe questions. However, PIP's applicability is limited to architectures employing explicit cross-attention mechanisms, and its reliance on question-conditioned attention constrains it exclusively to VQA tasks. Thus, the absence of a comprehensive and theoretically grounded investigation into the nature of AEs and adversarial detection for VLPs leaves a critical gap in understanding their vulnerabilities and robustness. 
\begin{figure*}[t]
\centering
\begin{subfigure}[t]{0.495\textwidth}
  \centering
  \includegraphics[width=\textwidth, trim=2 10 20 5, clip]{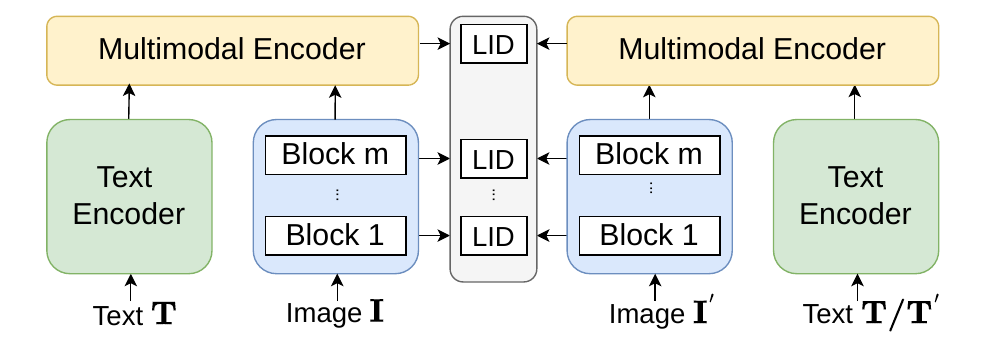}
  \caption{Extraction of LID scores.}
  \label{fig:pipeline1}
\end{subfigure}
\hfill
\begin{subfigure}[t]{0.495\textwidth}
  \centering
  \includegraphics[width=\textwidth, trim=2 10 20 5, clip]{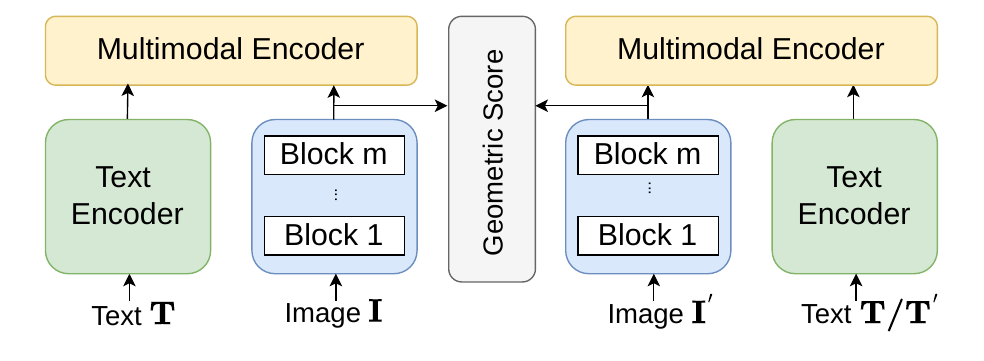}
  \caption{Extraction of $k$-NN, Mahalanobis, and KDE.}
  \label{fig:pipeline2}
\end{subfigure}
\caption{Pipeline of geometric score extraction for GeoDetect.}
\label{fig:pipeline}
\end{figure*}
In this work, we investigate the intrinsic properties of VLPs by revisiting the anisotropic nature of CLIP’s embedding space, as observed in prior work \cite{liang2022mind, levi2024double}, and extending this analysis to other VLPs. While existing studies primarily focus on CLIP, we show that this property extends to a broader range of VLPs, forming the foundation for our theoretical assumptions. 
Anisotropy indicates that representations are unevenly dispersed across embedding dimensions, creating dense and sparse directions. We leverage this property to formulate our central theoretical contribution: \textit{AEs explore off-manifold regions of the embedding space, resulting in a greater expected distance between an AE and a random clean example, compared to the distance between the unperturbed version of the same random clean example.} 
This motivates us to investigate the geometric properties surrounding data representations, through which we uncover fundamental differences between the regions occupied by adversarial and clean examples.

Building on our theoretical insights, we propose GeoDetect, an effective method for detecting AEs in VLPs. GeoDetect extracts deep representations from VLP encoders and applies classical geometric metrics to compute detection scores, including Local Intrinsic Dimensionality (LID) \cite{houle2013dimensionality}, k-Nearest Neighbours distance (k-NN), Mahalanobis distance~\cite{mclachlan1999mahalanobis}, and Kernel Density Estimation (KDE). 
These scores are then used for logistic regression or threshold-based detection of AEs. An overview of GeoDetect is illustrated in \cref{fig:pipeline}. \cref{fig:pipeline1} presents 
adversarial image detection using LID, while \cref{fig:pipeline2} illustrates detection using the other three studied methods, $k$-NN, Mahalanobis distance, and KDE. The key difference is that LID operates layer-wise, evaluating the outputs of both multimodal layers and other intermediate layers, while the other three methods operate on the output
of the image encoder, making LID more sensitive to perturbations across the multimodal encoder.
Built upon solid theoretical foundations, GeoDetect offers a cost-effective and robust method for ensuring the safety of VLPs.

The main contributions can be summarized as follows:
\begin{itemize}
    \item We analyze the anisotropic structure of VLP embedding spaces and, building on this, theoretically demonstrate that AEs tend to lie in off-manifold regions, resulting in larger geometric deviations from clean reference points.
    
    \item We introduce GeoDetect, a novel model-agnostic detection method that employs geometric discrepancies through a family of metrics to identify AEs in VLPs across different downstream tasks (e.g., zero-shot classification and retrieval).

    \item 
    We comprehensively validate the effectiveness of GeoDetect across various VLP architectures and attacks, achieving consistently high AUC scores, all without requiring fine-tuning. This makes GeoDetect both robust and lightweight compared to existing defense methods.

\end{itemize}

\section{Related Work}
\subsubsection{Vision-Language Pre-Trained Models.}
Vision-language representation learning outperforms visual representation learning across a wide range of tasks. 
For instance, CLIP uses a contrastive objective (i.e., InfoNCE loss \cite{oord2018representationInfoloss}) to align an image with its corresponding textual description in the feature space. 
VLPs aim to improve multimodal task performance by pre-training on large-scale image-to-text pairs \cite{li2022bliptozihvlp}. Several recent methods utilize pre-trained object detectors with region features as a foundation for obtaining vision-language representations \cite{chen2020uniter}. There are two primary types of VLPs depending on their architectures: fused and aligned \cite{zhang2022towards}. Fused VLPs, such as ALBEF and TCL \cite{yang2022visionTCL}, utilize distinct unimodal encoders to handle token embeddings and visual characteristics separately. They subsequently employ a multimodal encoder to produce integrated multimodal embeddings by combining image and text embeddings. Conversely, aligned VLPs such as CLIP are composed solely of unimodal encoders that have separate embeddings for image and text modalities. This research specifically examines widely used architectures, including both fused and aligned.

\subsubsection{Adversarial Attacks and Robustness in VLPs.}

Adversarial attacks aim to deceive deep learning models into misclassifying an input \cite{szegedy2013intriguing}.
While previous work is centered around image classification, recent studies show that VLPs are also vulnerable to adversarial attacks. 
For example, \cite{xu2018fooling} investigated attacks on visual question-answering models by altering the image modality. 
\cite{agrawal2018don,shah2019cycle} focused on disrupting vision-language models through text modality perturbations. \cite{zhang2022towards} offered key insights into the development of multimodal attacks and improving model robustness by exploring VLPs. Building on this, \cite{lu2023set, he2023sa,han2023ot} worked on enhancing the transferability of multimodal AEs by leveraging cross-modal interactions, data augmentation, and optimal transport theory. Furthermore, \cite{yin2023vlattack,zhou2023advclip} build on \cite{zhang2022towards} by crafting modality-aligned perturbations that improve transferability between downstream tasks. Despite these advances, many attack techniques remain specialized for classification tasks and may not generalize well to retrieval, captioning, or grounding. Therefore, we adopt the adversarial attacks presented in \cite{zhang2022towards, lu2023set} as our attack baselines. 

Recent efforts have explored enhancing the adversarial robustness of vision-language models through prompt tuning and training strategies. \cite{zhang2024adversarial,li2024one} improve CLIP’s resilience by learning robust textual prompts aligned with adversarial image embeddings. 
\cite{zhou2024few} introduces adversarial text supervision to balance cross-modal alignment and uni-modal discrimination. 
\cite{wang2024pre} adds an auxiliary branch to align adversarial outputs between the target and pre-trained models, reducing overfitting in zero-shot settings. \cite{wang2024revisiting} adopts a two-phase adversarial training regime, starting with lightweight pre-training, followed by high-resolution fine-tuning. However, the high computational overhead of these methods poses challenges for scaling to large models and datasets.
\subsubsection{Geometric Methods and Geometric Adversarial Detection in Unimodal Models.}
$k$-NN \cite{cover1967nearest} is a nonparametric algorithm that classifies points based on the majority label of their nearest neighbors, offering a simple yet powerful method for pattern recognition and regression. LID models the intrinsic dimensionality \cite{karger2002findingdimensionalmodeling, houle2012generalizeddimensionalmodeling, houle2013dimensionality, houle2017local, 
amsaleg2015estimating} near a point by analyzing the growth rate of nearby data, providing insights into local geometric structures within a dataset. Mahalanobis distance \cite{mclachlan1999mahalanobis} incorporates data covariance to measure similarity, enabling scale-invariant and correlation-sensitive evaluations that are effective for identifying outliers or understanding feature relationships. KDE estimates the probability density function of data in a nonparametric manner, using kernel functions and adaptive bandwidths to achieve smooth and flexible density representations \cite{botev2010kernel}.

Several studies have employed geometric approaches to detect AEs in unimodal classification models. \cite{grosse2017statistical} introduced the Maximum Mean Discrepancy (MMD), a kernel-based statistical test that distinguishes AEs from a model’s training data. 
As an alternative to KDE, \cite{ma2018characterizing} employed LID to evaluate the distance distribution of an input relative to its neighbors, capturing the local complexity of the sample’s surrounding space. \cite{lee2018simple} proposed using the Mahalanobis distance, using Gaussian discriminant analysis to detect out-of-distribution and adversarial samples through a generative classifier, offering a more refined confidence score than the traditional softmax classifier. \cite{cohen2020detecting} further explored $k$-NN for adversarial detection. While these methods have shown promise in unimodal settings, their effectiveness in VLPs remains unexplored.

\section{GeoDetect}
\label{Geoexplain}
In this section, we introduce GeoDetect, a geometric framework for detecting AEs by analyzing the 
properties of embedding geometry in VLPs. 
We first provide the formal problem definition in \cref{sec:details}, which is followed by a theoretical analysis in \cref{theoryanalysis}. 
\subsection{GeoDetect Framework}
\label{sec:details}

\subsubsection{Problem Setup.} Let $\mathcal{D}_c = \{(x_i, t_i)\}_{i=1}^{N}$ be a clean dataset of $N$ i.i.d. image-text pairs, where $x_i$ and $t_i$ denote the clean image and text, respectively.
The adversarial dataset consists of perturbed samples and is defined as $\mathcal{D}_a = \{(x_i', t_i)\}_{i=1}^{N}$ when only the image is perturbed, or $\mathcal{D}_a = \{(x_i', t_i')\}_{i=1}^{N}$ when both modalities are perturbed, with labels $y_i\!\in\!\{0,1\}$ indicating benign $(y_i=0)$ or adversarial $(y_i=1)$.
We define the embeddings as \( z_\mathrm{I} = E_\mathrm{I}(x) \) for image, \( z_\mathrm{T} = E_\mathrm{T}(t) \) for text, and in the case of fused VLPs, \( z_\mathrm{M} = E_\mathrm{M}(z_\mathrm{I},z_\mathrm{T}) \) as the multimodal representation. 
Clean embeddings are denoted as \( Z_c = \{z_i\}_{i=1}^N \), where \( z_i \) represents either \( z_{\mathrm{I}} \) or \( z_{\mathrm{M}} \). Our goal is to accurately detect perturbed samples, particularly those in which the image, or both the image and text modalities, have been adversarially modified. Given a query $(x_i,t_i)$ and a reference batch $\{(x_j,t_j)\}_{j=1}^{n}$, with \( n \) denoting the batch size, we evaluate the detection function:
\begin{equation}
\label{metricdetector}
    f((x_i,t_i), (x_j,t_j)_{j=1}^{n}) = 
    \mathcal{H}(\text{Metric}(z_i, \{z_j\}_{j=1}^{n})), 
\end{equation}
where \( \{z_j\}_{j=1}^{n, j \neq i} \) denotes a set of clean reference embeddings, \( n \) is the batch size, 
and 
$\mathcal{H}$ represents the decision function. 
The function \( \text{Metric}(\cdot, \cdot) \) represents a geometric measure, such as LID, \( k \)-NN, KDE, or Mahalanobis distance. In this paper, we adopt the maximum likelihood estimation (MLE) of LID \cite{amsaleg2015estimating}, and throughout the paper, we use "LID" to refer to this estimated quantity. The binary classification \( f(\cdot, \cdot) \) then determines whether the input is adversarial or clean using the computed score.

\subsubsection{GeoDetect Pipeline.} GeoDetect comprises three primary steps: generation, extraction, and detection. Following prior work~\cite{ma2018characterizing}, we assume that the defender has access to a subset of the data, and that the initial dataset $\mathcal{D}_c$ is free of AEs.

In the first step of the process, generation, AEs are created from clean samples using different adversarial attacks. Given the clean dataset, we generate perturbed image–text pairs $(x^{'}_{i}$ and $t^{'}_{i})$, resulting in a balanced set with equal proportions of clean and adversarial samples. A description of adversarial‐sample generation, including the settings for each attack type, is provided in Appendix A.1.

In the extraction step, we begin by extracting clean image embeddings, $z_\mathrm{I}$, and multimodal embeddings, $z_\mathrm{M}$, as well as the corresponding adversarial embeddings $z'_\mathrm{I}$ and $z'_\mathrm{M}$. To ensure scalability on large datasets, we use minibatch sampling to estimate local geometric properties, following the approach in \cite{ma2018characterizing}, which has been shown to provide reliable approximations of neighborhood statistics.
To compute geometric scores of a target embedding $z_i$, we randomly sample a batch of clean embeddings \(\{z_j\}_{j=1}^{n}\) as reference points for the computation of different metrics. These reference points can be either $z_{\mathrm{I}}$ or $z_{\mathrm{M}}$, depending on whether the detection is performed in the image or multimodal space.  Using this reference batch, scores will be computed for both clean $z_i$ and adversarial embeddings $z'_{i}$ using the following metrics as $\mathrm{Metric}(\cdot,\cdot)$:

{\small
\[
\mathrm{Metric}(\cdot,\cdot)=
\left\{
\begin{array}{l}
\mathrm{k\text{-}NN}(z_i,\{z_j\})=\frac1k\sum_{j=1}^k r_j(z_i),
\; r_j=\|z_i-z_j\|_2,\\
\widehat{\mathrm{LID}}(z_i,\{z_j\})
=\!\left(-\frac1k\sum_{j=1}^k
\log\frac{r_j}{r_{\max}}\right)^{-1},\\
\mathrm{KDE}(z_i,\{z_j\};H)=\frac1n\sum_{j=1}^n K_H(z_i-z_j),\\
\mathrm{Mahal}(z_i)=
\sqrt{(z_i-\mu)^T\Sigma^{-1}(z_i-\mu)}.
\end{array}
\right.
\]
}

For the Mahalanobis distance, the mean vector \( \mu \) and covariance matrix \( \Sigma \) are computed using the clean dataset \( \mathcal{D}_c \). Similarly, for KDE, the kernel function \( K_{H}(\cdot, \cdot) \) is estimated based on the same clean data. For all metrics, the embedding-level scores are computed as \( s_i = \text{Metric}(z_i, \{z_j\}_{j=1}^{n}) \) for clean samples, and \( s'_i = \text{Metric}(z'_i, \{z_j\}_{j=1}^{n}) \) for adversarial samples, where the reference points $\{z_j\}_{j=1}^{n}$ are clean embedding samples. For LID, we follow a layer-wise extraction strategy as proposed by \cite{ma2018characterizing}. In fused VLPs, we additionally compute the LID of the multimodal encoder \(z_\mathrm{M}\) as an additional feature alongside the image encoder layers, improving detection performance against multimodal attacks. The complete procedure for computing these values for adversarial image detection is outlined in Algorithm 1 
in Appendix A.2. 
After the extraction step, the extracted clean and adversarial scores are denoted as \( s_i \in S_{(N,l)} \) and \( s'_i \in S'_{(N,l)} \), where \( l=1 \) for \( k \)-NN, Mahalanobis, and KDE, and \( l \) represents the number of layers for LID. These extracted scores serve as input features for the subsequent detection phase.

We frame adversarial example detection via a decision function $\mathcal{H}$.
We split the extracted scores into a calibration subset (to set thresholds or train a classifier) and a test subset.
For $k$-NN, Mahalanobis, and KDE, $\mathcal{H}$ is a threshold rule on the scalar score
$s_i=\mathrm{Metric}\!\big(z_i,\{z_j\}_{j=1}^{n}\big)$:
\(
\mathcal{H}(s_i)=\mathbb{I}(s_i>\tau),
\)
where $\tau$ is chosen on the training split.
For LID, $\mathcal{H}$ is a logistic regression trained on multi-layer LID features (Appendix A.2
). At test time, given $(x_i,t_i)$, we extract embeddings $(z_{\mathrm I},z_{\mathrm T})$, compute the chosen metric with respect to a clean reference batch $\{z_j\}_{j=1}^{n}$, and apply $\mathcal{H}$ (threshold or logistic) to decide if the input is adversarial.
The efficiency of GeoDetect is reported in Appendix A.4
.

\subsection{GeoDetect Theoretical Analysis}
\label{theoryanalysis}
In this subsection, we develop GeoDetect’s theoretical foundations by formalizing core assumptions and deriving its key technical results.

We begin by empirically verifying that VLP embeddings are anisotropic, concentrated along a few dominant directions, 
which motivates our assumptions. Building on this, we show that the principal directions of adversarial embeddings differ significantly from those of clean embeddings, effectively pushing them off the manifold. 
Our theoretical analysis explains why geometric scores are particularly well-suited for adversarial detection in VLPs: they quantify off-manifold deviations that adversarial perturbations inherently induce.

\begin{figure}[t]
    \centering

    \makebox[\textwidth][c]{%
    \begin{subfigure}[t]{0.68\textwidth}
        \centering
        \begin{tikzpicture}
        \begin{groupplot}[
            group style={group size=2 by 1, horizontal sep=0.85cm},
            width=0.55\textwidth,
            height=3.2cm,
            grid=major,
            tick label style={font=\scriptsize},
            label style={font=\scriptsize},
            title style={font=\scriptsize},
            x label style={font=\scriptsize},
            y label style={font=\scriptsize},
            xlabel={Iteration},
            ylabel={Score},
            xlabel near ticks,
            ylabel near ticks,
        ]
        \nextgroupplot[
            title={$I_1$ Score},
            legend style={
                at={(1.0,-0.45)},
                anchor=north,
                draw=none,
                fill=none,
                legend columns=2,
                column sep=3pt,
                row sep=0pt,
                font=\scriptsize,
                nodes={scale=0.9, transform shape},
            },
            legend cell align=left,
            legend image post style={xscale=0.55},
        ]
            \addplot[blue, mark=*, mark size=1pt]
            table[x index=0, y index=1, col sep=space] {files/i1resnet.dat};
            \addlegendentry{Supervised}
            \addplot[red, mark=*, mark size=1pt]
            table[x index=0, y index=1, col sep=space] {files/i1albef.dat};
            \addlegendentry{ALBEF}

        \nextgroupplot[
            title={$I_2$ Score},
            ymin=-0.0002, ymax=0.0025,
        ]
            \addplot[blue, mark=*, mark size=1pt]
            table[x index=0, y index=1, col sep=space] {files/i2resnet.dat};
            \addplot[red, mark=*, mark size=1pt]
            table[x index=0, y index=1, col sep=space] {files/i2albef.dat};
        \end{groupplot}
        \end{tikzpicture}
        \caption{Isotropy scores ($I_1$, $I_2$) across iterations.}
        \label{fig:isotropy}
    \end{subfigure}
    \hspace{0.015\textwidth}
    \begin{subfigure}[t]{0.35\textwidth}
        \centering
        \begin{tikzpicture}
        \begin{axis}[
            width=\textwidth,
            height=3.2cm,
            ybar,
            bar width=9pt,
            ymin=0, ymax=60,
            tick label style={font=\scriptsize},
            label style={font=\scriptsize},
            ylabel style={font=\scriptsize},
            title style={font=\scriptsize},
            ylabel={$\widehat{\mathrm{ER}}$},
            symbolic x coords={CLIP$_{\mathrm{CNN}}$, CLIP$_{\mathrm{ViT}}$, ALBEF, TCL, Sup.},
            xtick=data,
            xticklabel style={rotate=40, anchor=east, font=\scriptsize},
        ]
            \addplot coordinates {
                (CLIP$_{\mathrm{CNN}}$, 20.1)
                (CLIP$_{\mathrm{ViT}}$, 28.6)
                (ALBEF, 24.7)
                (TCL, 20.2)
                (Sup., 57.8)
            };
        \end{axis}
        \end{tikzpicture}
        \caption{Normalized ER comparison.}
        \label{fig:effective}
    \end{subfigure}%
    }

    \caption{Comparison of isotropy metrics and effective ranking for VLPs vs. supervised model. Iteration refers to the batch index during evaluation.}
\end{figure}
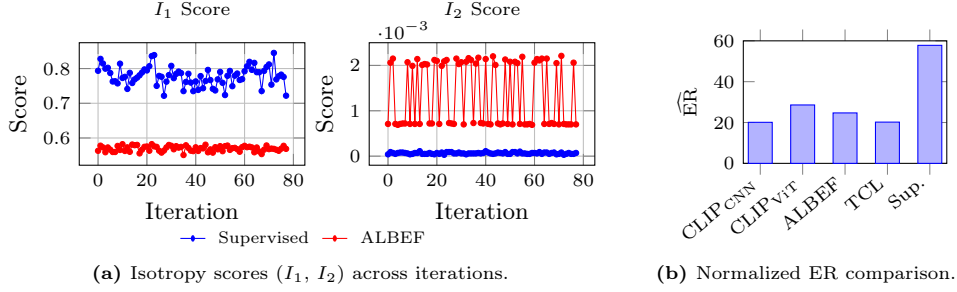

\subsubsection{Anisotropic Embedding Space.}
\label{isotropic}

Understanding the geometric structure of the embedding space is crucial for identifying the fundamental differences between clean and adversarial embeddings in VLPs. \cite{liang2022mind} showed that CLIP features lie within a low-aperture cone, concentrating 
in a narrow angular region of high-dimensional space. Building on this, \cite{levi2024double} found that CLIP’s embedding space forms a double-ellipsoid geometry, with image and text embeddings located on distinct ellipsoidal shells. 

Consequently, we expect that other VLPs also exhibit anisotropic embedding spaces, consistent with the patterns observed in CLIP. To formally quantify this, we adopt two established isotropy measures, \( I_1 \) and \( I_2 \) 
\cite{wang2019improvingisotropic}. Let $Z \in R ^{N \times D}$ be the matrix of embedding vectors. The measures are defined as: 
\begin{equation}
I_1(Z) = \frac{\min_{v \in V} P(v)}{\max_{v \in V} P(v)}, \quad I_2(Z) = \sqrt{\frac{\sum_{v \in V} \left( P(v) - \bar{P}(v) \right)^2}{|V| \bar{P}(v)^2}},
\end{equation}
where $V$ is the set of eigenvectors of $Z^T Z$, and $P: \mathbb{R}^D \to \mathbb{R}^+$ is the partition function $P(v) = \sum_{i=1}^n \exp(\langle v, z_i \rangle)$ \cite{mu2018all}. For an embedding matrix \( Z \) to be isotropic, $P(v)$ should be approximately constant for any unit vector $v$ \cite{arora2016latent}. Then, the second measure, $I_2(Z)$, is the normalized standard deviation of the partition function $P(v)$, where $\bar{P}(v)$ is the average value of $P(v)$. In this formulation, $I_1(Z)$ is bounded between 0 and 1 ($I_1(Z) \in [0, 1]$), while $I_2(Z)$ is non-negative ($I_2(Z) \geq 0 $)
; lower $I_1(Z)$ and higher $I_2(Z)$ indicate stronger anisotropy.

To investigate the embedding space of the VLPs (specifically ALBEF here, with other VLPs discussed in Appendix B.1
), we empirically evaluate $I_1(Z)$ and $I_2(Z)$. As a reference, we include a supervised classifier (ResNet-50) trained on ImageNet. 
We compute \(I_1(Z)\) and \(I_2(Z)\) across iterations (x-axis) and report the corresponding metrics on the y-axis in \cref{fig:isotropy}.
The results show that \(I_1(Z)\) is lower and \(I_2(Z)\) is higher for VLPs, shown here for ALBEF, compared to the supervised classifier, indicating stronger anisotropy in VLP embedding spaces. Consistent trends for other VLPs are reported in Appendix B.1
, supporting the assumptions underlying our theoretical analysis. 

Another metric for evaluation of isotropy is effective rank (ER), a spectral measure of dimensionality that reflects how singular values are distributed, providing a more complex perspective compared to the traditional rank \cite{roy2007effective}. Mathematically, the ER of a matrix \(Z\) 
is defined based on the spectral entropy of its normalized singular values. 
Let \(\sigma_i\) be the singular values of \(Z\), and \( \hat{\sigma_i} \) represent the normalized singular values \( \displaystyle \hat{\sigma_i} = \frac{\sigma_i}{\sum_j \sigma_j} \), where \(\sum_i \hat{\sigma_i} = 1\). The spectral entropy and the normalized ER, denoted as \( \widehat{\text{ER}} \) (scaled by the logarithm of the dimension $D$), are given by:
\begin{equation}
H = -\sum_i \hat{\sigma_i} \log \hat{\sigma_i} \Rightarrow \widehat{\text{ER}}(Z) = \frac{\exp(H)}{\log{D}}.
\end{equation}
In the isotropic setting, where all singular values are equal, the estimated effective rank \( \widehat{\text{ER}} \) approaches the full rank of the matrix. In contrast, when the singular values are concentrated in a few dimensions, \( \widehat{\text{ER}} \) is substantially lower, indicating anisotropy. \cref{fig:effective} compares the normalized effective rank of various VLPs with that of a supervised ResNet-50 trained on ImageNet V2. The results show that VLPs consistently exhibit lower \( \widehat{\text{ER}} \), confirming that their embedding spaces are more anisotropic than those of supervised models.

\subsubsection{Adversarial Perturbation Effect on Geometry.}

In this subsection, we establish the theoretical assumptions and provide key analyses underpinning our geometric-based detection methodology. 
Let $\Sigma \in R^{D \times D}$ denote the covariance matrix of the clean embedding distribution. We denote clean embeddings as $z_i \in Z_c$
, and adversarial embeddings as $z'_i \in Z_a$. We further represent the distributions of clean and adversarial embeddings by $p(z)$ and $q(z')$, respectively. 
To characterize the geometric deviation induced by adversarial perturbations, we define the distance of a target embedding (clean or adversarial) to a randomly sampled clean embedding $z_u$, where $u \neq i$, as: 
\[Dist_{\text{c}}=\|z_i - z_u\| \quad \text{and} \quad Dist_{\text{a}}=\|z_i' - z_u\|.\] 
We analyze the expected distances $\mathbb{E}[Dist_{\text{c}}]$ and $\mathbb{E}[Dist_{\text{a}}]$, and formally demonstrate that adversarial embeddings have higher expected distances, an observation underlying the effectiveness of geometric metrics in adversarial detection. We first state two assumptions that ground our theoretical analysis:

\begin{assumption}[Anisotropic Covariance]\label{ass:anisotropic_covariance} 
The covariance matrix $\Sigma \in \mathbb{R}^{D \times D}$ of the clean embedding space is positive-definite and anisotropic, specifically $\Sigma \neq c I$ for any scalar constant $c$, indicating anisotropy property. Consequently, its eigenvalues vary significantly across dimensions ($\sigma_1 \gg \sigma_2 \gg ... \gg \sigma_D$).
\end{assumption}

\begin{assumption}[Manifold Proximity]\label{ass:gaussian} 
Clean embeddings reside on a manifold  \( \mathcal{M} \), such that the distance of a data point \( z_i \) from the manifold satisfies a proximity condition \( \|z_i - \mathcal{M}\| \leq \alpha \). Additionally, given the high dimensionality of the embeddings, we assume Gaussian distributions for both clean and adversarial embeddings:
\(
p(z) \sim \mathcal{N}(\mu_z, \Sigma), \quad q(z') \sim \mathcal{N}(\mu_{z'}, \Sigma').
\)
\end{assumption}
\begin{figure}[t]
    \centering
    
    \begin{subfigure}[b]{0.3\textwidth}
        \includegraphics[width=\textwidth]{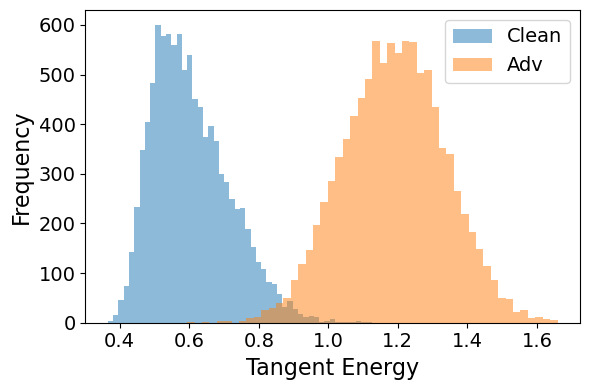}
        \caption{Comparison of energy in normal direction.}
        \label{fig:off1}
    \end{subfigure}
    \hfill
    \begin{subfigure}[b]{0.3\textwidth}
        \includegraphics[width=\textwidth]{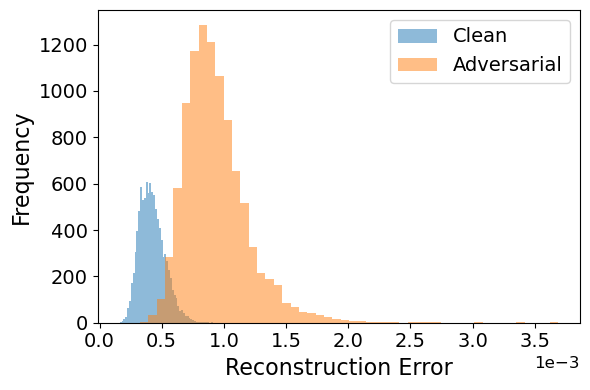}
        \caption{Comparison of reconstruction error.}
        \label{fig:off2}
    \end{subfigure}
    \hfill
    \begin{subfigure}[b]{0.3\textwidth}
        \includegraphics[width=\textwidth]{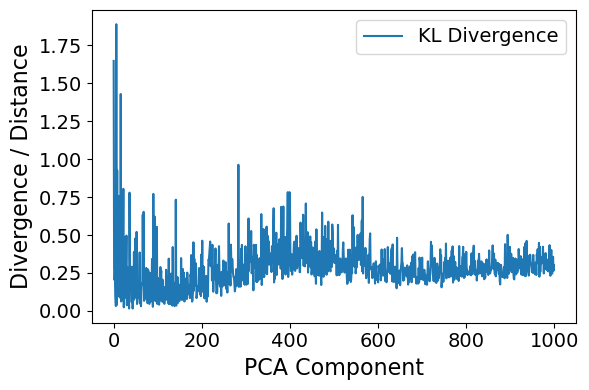}
        \caption{Divergence per PCA dimension (ordered by variance).}
        \label{fig:off3}
    \end{subfigure}
    
    \caption{Verification of Lemma \ref{lemma:offmanifold}: adversarial data in VLPs are off-manifold.}
    \label{fig:all}
\end{figure}

Assumption \ref{ass:anisotropic_covariance}, verified in \cref{isotropic}, explains the anisotropic geometry observed in VLP embeddings. Assumption \ref{ass:gaussian}, built on the manifold hypothesis \cite{bengio2013representation}, distinguishes clean embeddings that lie near the manifold from adversarial embeddings that deviate from it. To justify Assumption \ref{ass:gaussian}, we note that although data may be globally non-Gaussian or manifold-valued, it is standard to analyze them through local neighborhoods: non-Gaussian structures often appear approximately Gaussian when viewed locally, and curved manifolds can be locally approximated by Euclidean spaces \cite{lee2006riemannian}. As emphasized by \cite{zhao2007laplacian}, even globally non-Gaussian or manifold-valued data exhibit locally Gaussian behavior, since any curved manifold is locally Euclidean. Building on these assumptions, and inspired by Theorem 1 in \cite{zhang2024goal}, we derive the optimal adversarial embedding. 
\begin{lemma}
\label{lemma:optimum}

Following Assumption \ref{ass:gaussian}, let clean and adversarial embeddings follow \( p(z) \sim \mathcal{N}(\mu_z, \Sigma)\) and \(q(z') \sim \mathcal{N}(\mu_{z'}, \Sigma'),\) respectively, where $\Sigma$ is a fixed positive-definite covariance. Maximizing the KL divergence $
\mathrm{KL}\bigl(q\big\|p\bigr)$ is approximately equivalent to maximizing the quadratic form of $(z_i' - z_i)^\top \Sigma^{-1} (z_i' - z_i)$, which can be
transformed into a Lagrangian minimization optimization
problem, which has an optimal closed-form solution:
\begin{equation}
    z_i'{}^{*} = (\Sigma + \lambda I)^{-1} \lambda z_i, \quad \lambda > 0,
\end{equation}
where \( \lambda \) is the Lagrange multiplier.

\end{lemma}
The proof of Lemma \ref{lemma:optimum} appears in Appendix B.2
. 

\begin{lemma} \label{lemma:offmanifold} 

Following Assumption \ref{ass:anisotropic_covariance} and Lemma \ref{lemma:optimum}, 
let $\mathcal{M}$ represent the data manifold formed by clean embeddings within a local batch. 
When the data is perturbed, and assuming that clean embeddings lie close to the manifold $\mathcal{M}$, the resulting embeddings deviate from the manifold $\mathcal{M}$, thus characterized as off-manifold. Specifically, given the optimal adversarial embedding $
z_i^{\prime *} = (\Sigma + \lambda I)^{-1} \lambda z_i, \quad \lambda > 0,
$ the deviation from the manifold satisfies $\|z_i^{\prime} - \mathcal{M}\| \geq \gamma$, where $\gamma > \alpha \geq 0$ defines the minimum separation threshold for off-manifold data.
\end{lemma}

Lemma~\ref{lemma:offmanifold} shows that adversarial embeddings leave the clean manifold by suppressing tangent components and amplifying normal components (proof in Appendix B.3
). We empirically verify that adversarial perturbations move samples off the clean manifold using ImageNet-V2 and CLIP\textsubscript{CNN}. 
\cref{fig:off1} shows that adversarial embeddings concentrate more energy in the lowest-singular-value directions (rarely used by clean data), consistent with motion into manifold-normal space. 
\cref{fig:off2} further shows higher low-rank PCA reconstruction residuals for adversarial samples (top $K{=}10$ clean PCs), indicating increased distance from the clean low-dimensional subspace. 
Finally, \cref{fig:off3} reports larger KL divergence between clean and adversarial distributions along leading PCA axes (most pronounced in the top 1--20 components), confirming a significant shift in the most informative representation directions.

\begin{theorem}[Expected Distance Gap]\label{theorem:distance}
Following Lemmas \ref{lemma:optimum} and \ref{lemma:offmanifold}, let 
\(
z_i \sim p(\cdot)
\)
be a clean embedding satisfying \(\|z_i - \mathcal{M}\|\le \alpha\), and let 
\(
z_i' 
\)
be an adversarial embedding satisfying \(\|z_i' - \mathcal{M}\|\ge \gamma\) with \(\gamma > 3\alpha\).  Then
\begin{equation}
    \mathbb{E}_{z_u\sim p(\cdot)}\bigl[\|z_i' - z_u\|\bigr]
>
\mathbb{E}_{z_u\sim p(\cdot)}\bigl[\|z_i - z_u\|\bigr].
\end{equation}

\end{theorem}

Theorem~\ref{theorem:distance} implies that given a query embedding, a large distance to randomly sampled clean embeddings strongly indicates adversarial perturbation. This justifies the effectiveness of geometric-based metrics for adversarial detection. Proof of Theorem~\ref{theorem:distance} is included in Appendix B.4
, and details on the connection between Theorem~\ref{theorem:distance} and Lemma~\ref{lemma:offmanifold} to different geometric-based approaches are provided in Appendix B.5
. Specifically, we demonstrate that under mild assumptions, AEs are expected to exhibit higher LID, $k$-NN, and Mahalanobis scores, along with lower KDE scores. Empirical evidence supporting Lemma~\ref{lemma:offmanifold} and Theorem~\ref{theorem:distance} is provided in Appendix B.6.

\section{Experiments}
\label{experiment}

We evaluate GeoDetect on standard VLP tasks, including zero-shot classification and image-text retrieval.
We compare against MCM \cite{ming2022delving}, which is designed for CLIP-based classification and detects out-of-distribution inputs via softmax-normalized similarity scores. MCM is a score-based zero-shot baseline that can be directly applied to AE detection in multimodal models.
 While relevant for classification-based VLP settings, its reliance on discrete class labels makes it incompatible with retrieval tasks.

\subsection{Experimental Setup}
\subsubsection{Datasets and Models.}
We evaluate zero-shot classification with ImageNet \cite{deng2009imagenet}, CIFAR10, CIFAR100 \cite{krizhevsky2009cifar}, STL-10 \cite{coates2011analysis}, and Food-101 \cite{bossard2014food}, as the standard datasets for zero-shot classification. Following \cite{radford2021clip}, we use class prompts of the form "a photo of a $c$"
, where $c$ is the name of the class. For image-text retrieval, we conduct experiments on commonly used datasets, 
Flickr30K \cite{young2014flickr} and MS-COCO \cite{lin2014microsoft}. We consider two types of VLPs: aligned and fused. For aligned VLPs, we evaluate CLIP\textsubscript{ViT} (using ViT-B/16) and CLIP\textsubscript{CNN} (using ResNet-50) \cite{radford2021clip}. For fused VLPs, we examine ALBEF \cite{li2021align} and TCL \cite{yang2022visionTCL}, which consist of separate image, text, and multimodal encoders. For image-text retrieval experiments on MSCOCO and Flickr30k, we use the publicly released fine-tuned checkpoints of ALBEF and TCL.

\subsubsection{Threat Models and Evaluation Metrics.}
We follow Sep-Attack and Co-Attack methods \cite{zhang2022towards} due to their applicability to different models and tasks. 
The [CLS] embedding is widely used in pre-trained models for downstream tasks; therefore, we use it as the attack target in our evaluation. Sep-Attack perturbs each modality independently, while Co-Attack jointly targets both modalities. For image-focused attacks, we evaluate two variants of Sep-Attack: Sep\textsubscript{uni}, which targets unimodal embeddings, and Sep\textsubscript{multi}, which targets the fused multimodal representation (applicable only to fused VLPs). For image attacks, consistent with \cite{zhang2022towards}, we adopt iterative adversarial 
perturbations constrained in the $\ell_\infty$ norm, and a BERT-style \cite{li2020bert} attack strategy for text attack. 
The maximum perturbation $\epsilon_{i}$ is set to $8/255$, with a step size of $1.25$ for 10 iterations.  For text, the perturbation budget is set to 1 token. We also evaluated the SGA attack \cite{lu2023set}, an improved version of Co-Attack, in Appendix D.2. Detailed attack configurations and success rates are reported in Appendix A.1 
and Appendix A.5
, respectively. We assess performance using two standard metrics: (1) the false positive rate at 95\% true positive rate (FPR95), and (2) the area under the receiver operating characteristic curve (AUC).

\subsubsection{Settings and Layers.}
 
For CLIP\textsubscript{CNN} and CLIP\textsubscript{ViT}, we use batch size $128$ with $k{=}100$ for LID, $k{=}10$ for $k$-NN, and a Gaussian KDE bandwidth of $0.1$. For ALBEF and TCL, batch size is 64 with $k{=}40$ for LID, $k{=}10$ for $k$-NN, and the same KDE bandwidth. Adversarial examples are generated from the entire test set. The resulting mixed dataset (clean and adversarial) is then randomly split into 80\% for calibration (threshold fitting / LID training) and 20\% for evaluation. Detailed layer selection is provided in Appendix A.3
, with layer sensitivity in Appendix C.3
. \footnote{The code is publicly available in \url{https://github.com/AfsanehEB/GeoDetect}.}

\subsection{GeoDetect Performance}
\begin{table*}[t]
\centering
\caption{
Results on zero-shot classification performance using the area under the receiver operating characteristic curve (AUC) and the false positive rate at 95\% true positive rate (FPR95). Higher AUC ($\uparrow$) and lower FPR95 ($\downarrow$) values indicate more accurate detection.
}
\label{res_combined1}
\small 
\begin{adjustbox}{width=0.9\linewidth}
\begin{tabular}{@{}ccccccccccccc@{}}
\toprule
\multirow{2}{*}{\textbf{Model}} & \multirow{2}{*}{\textbf{Method}} & \multirow{2}{*}{\textbf{Attack}} & \multicolumn{2}{c}{\textbf{CIFAR10}} & \multicolumn{2}{c}{\textbf{CIFAR100}} & \multicolumn{2}{c}{\textbf{ImageNet1k}} & \multicolumn{2}{c}{\textbf{STL10}} & \multicolumn{2}{c}{\textbf{Food101}} \\ \cmidrule(l){4-13} 
 &  &  & \textbf{AUC} & \textbf{FPR95} & \textbf{AUC} & \textbf{FPR95} & \textbf{AUC} & \textbf{FPR95} & \textbf{AUC} & \textbf{FPR95} & \textbf{AUC} & \textbf{FPR95} \\ \midrule
\multirow{10}{*}{CLIP\textsubscript{CNN}} & \multirow{2}{*}{MCM} & Sep\textsubscript{uni} & 65.47 & 82.88 & 41.13 & 94.15 & 86.10 & 60.35 & 95.82 & 17.92 & 91.70 & 40.18 \\
 &  & Co-Attack & 67.10 & 79.54 & 43.99 & 93.21 & 80.83 & 68.38 & 94.10 & 25.64 & 82.14 & 64.38 \\ \cmidrule(l){2-13} 
 & \multirow{2}{*}{\begin{tabular}[c]{@{}c@{}}GeoDet-\\ LID\end{tabular}} & Sep\textsubscript{uni} & 100 & 0.00 & 100 & 0.00 & 99.31 & 1.87 & 100 & 0.00 & 99.98 & 0.06 \\
 &  & Co-Attack & 100 & 0.00 & 100 & 0.00 & 99.50 & 1.62 & 100 & 0.00 & 99.95 & 0.08 \\ \cmidrule(l){2-13} 
 & \multirow{2}{*}{\begin{tabular}[c]{@{}c@{}}GeoDet-\\ $k$-NN\end{tabular}} & Sep\textsubscript{uni} & 100 & 0.00 & 100 & 0.00 & 99.65 & 1.62 & 100 & 0.00 & 100 & 0.00 \\
 &  & Co-Attack & 100 & 0.00 & 100 & 0.00 & 99.67 & 0.89 & 100 & 0.00 & 100 & 0.00 \\ \cmidrule(l){2-13} 
 & \multirow{2}{*}{\begin{tabular}[c]{@{}c@{}}GeoDet-\\ Mah.\end{tabular}} & Sep\textsubscript{uni} & 100 & 0.00 & 100 & 0.00 & 96.62 & 9.32 & 99.88 & 0.33 & 99.79 & 1.16 \\
 &  & Co-Attack & 100 & 0.00 & 100 & 0.00 & 97.28 & 7.97 & 99.80 & 0.59 & 99.38 & 2.32 \\ \cmidrule(l){2-13} 
 & \multirow{2}{*}{\begin{tabular}[c]{@{}c@{}}GeoDet-\\ KDE\end{tabular}} & Sep\textsubscript{uni} & 100 & 0.00 & 100 & 0.00 & 98.72 & 7.24 & 99.87 & 0.26 & 100 & 0.00 \\
 &  & Co-Attack & 100 & 0.00 & 100 & 0.00 & 99.33 & 2.81 & 99.85 & 0.33 & 100 & 0.00 \\ \midrule
\multirow{15}{*}{ALBEF} & \multirow{3}{*}{MCM} & Sep\textsubscript{uni} & 91.20 & 29.02 & 82.80 & 49.19 & 92.15 & 25.38 & 96.83 & 16.02 & 90.26 & 37.03 \\
 &  & Sep\textsubscript{multi} & 47.43 & 98.23 & 33.55 & 99.56 & 63.03 & 97.59 & 65.32 & 86.60 & 41.98 & 99.46 \\
 &  & Co-Attack & 93.34 & 24.64 & 82.67 & 47.27 & 92.14 & 24.94 & 96.45 & 19.70 & 81.29 & 74.70 \\ \cmidrule(l){2-13} 
 & \multirow{3}{*}{\begin{tabular}[c]{@{}c@{}}GeoDet-\\ LID\end{tabular}} & Sep\textsubscript{uni} & 100 & 0.00 & 99.97 & 0.05 & 91.85 & 29.41 & 99.64 & 1.62 & 99.87 & 0.67 \\
 &  & Sep\textsubscript{multi} & 99.96 & 0.20 & 99.85 & 0.44 & 78.77 & 67.68 & 96.63 & 15.65 & 92.31 & 33.27 \\
 &  & Co-Attack & 100 & 0.00 & 99.98 & 0.05 & 93.85 & 20.07 & 99.85 & 0.69 & 99.92 & 0.42 \\ \cmidrule(l){2-13} 
 & \multirow{3}{*}{\begin{tabular}[c]{@{}c@{}}GeoDet-\\ $k$-NN\end{tabular}} & Sep\textsubscript{uni} & 100 & 0.00 & 100 & 0.00 & 98.60 & 7.23 & 99.97 & 0.19 & 99.98 & 0.04 \\
 &  & Sep\textsubscript{multi} & 99.27 & 3.05 & 99.21 & 3.25 & 51.92 & 93.61 & 75.95 & 75.75 & 86.46 & 50.96 \\
 &  & Co-Attack & 100 & 0.00 & 100 & 0.00 & 98.64 & 7.33 & 99.96 & 0.19 & 99.98 & 0.04 \\ \cmidrule(l){2-13} 
 & \multirow{3}{*}{\begin{tabular}[c]{@{}c@{}}GeoDet-\\ Mah.\end{tabular}} & Sep\textsubscript{uni} & 100 & 0.00 & 100 & 0.00 & 99.94 & 0.20 & 100 & 0.00 & 100 & 0.00 \\
 &  & Sep\textsubscript{multi} & 100 & 0.00 & 100 & 0.00 & 81.41 & 64.82 & 99.25 & 3.19 & 99.16 & 3.92 \\
 &  & Co-Attack & 100 & 0.00 & 100 & 0.00 & 99.93 & 0.25 & 100 & 0.00 & 100 & 0.00 \\ \cmidrule(l){2-13} 
 & \multirow{3}{*}{\begin{tabular}[c]{@{}c@{}}GeoDet-\\ KDE\end{tabular}} & Sep\textsubscript{uni} & 99.38 & 0.71 & 100 & 0.00 & 96.78 & 16.93 & 99.70 & 1.06 & 99.95 & 0.16 \\
 &  & Sep\textsubscript{multi} & 99.24 & 0.86 & 99.85 & 0.81 & 66.83 & 81.75 & 88.63 & 66.19 & 87.61 & 49.27 \\
 &  & Co-Attack & 99.38 & 0.76 & 100 & 0.00 & 96.67 & 18.09 & 99.72 & 1.00 & 99.94 & 0.16 \\ \bottomrule
\end{tabular}
\end{adjustbox}
\end{table*}

\subsubsection{Performance of GeoDetect in Zero-Shot Classification.}
As shown in \cref{res_combined1}, our geometric approaches consistently outperform the MCM method across all datasets in CLIP\textsubscript{CNN}, achieving lower FPR and higher AUC. This highlights GeoDetect’s effectiveness for AE detection. Among the evaluated metrics, $k$-NN surpasses other metrics (particularly Mahalanobis and KDE) in CLIP\textsubscript{CNN}, with LID showing comparable performance to $k$-NN in this context. 
On ALBEF, Mahalanobis slightly outperforms other metrics, particularly KDE, highlighting its sensitivity to image-level perturbations, while LID performs comparably in multimodal attacks, emphasizing the value of incorporating multimodal embeddings.
Despite ALBEF’s multimodal design, image perturbations still yield detectable shifts in the embedding space, which Mahalanobis effectively captures by modeling the covariance of clean image features. Extended results for CLIP\textsubscript{ViT} and TCL in Appendix E.1 
exhibit consistent patterns with those in this subsection. 

\begin{table*}[t]
\caption{
Results on image-text retrieval with Flickr30k and COCO dataset evaluated using the area under the receiver operating characteristic curve (AUC) and the false positive rate at 95\% true positive rate (FPR95). Higher AUC ($\uparrow$) and lower FPR95 ($\downarrow$) values indicate more accurate detection.
}
\label{nonclassification}

\begingroup
\scriptsize
\setlength{\tabcolsep}{2.5pt}        
\renewcommand{\arraystretch}{1.05}    

\begin{subtable}[t]{.47\linewidth}
\centering
\caption{Results for CLIP\textsubscript{CNN} and ALBEF Models}

\resizebox{\linewidth}{!}{%
\begin{tabular}{@{}lllcccc@{}}  
\toprule
\multirow{3}{*}{\textbf{Model}} &
\multirow{3}{*}{\textbf{Method}} &
\multirow{3}{*}{\textbf{Attack}} &
\multicolumn{4}{c}{\textbf{Dataset}} \\
\cmidrule(lr){4-7}
& & & \multicolumn{2}{c}{Flickr30k} & \multicolumn{2}{c}{COCO} \\
\cmidrule(lr){4-5}\cmidrule(lr){6-7}
& & & AUC & FPR95 & AUC & FPR95 \\
\midrule

\multirow{4}{*}{CLIP\textsubscript{CNN}} &
\multirow{2}{*}{LID} &
Sep\textsubscript{uni} & 98.45 & 4.52 & 99.54 & 1.46 \\
& & Co-Attack         & 98.90 & 4.52 & 99.50 & 1.56 \\
\cmidrule{2-7}
& \multirow{2}{*}{$k$-NN} &
Sep\textsubscript{uni} & 99.99 & 0.00 & 99.97 & 0.00 \\
& & Co-Attack         & 99.97 & 0.00 & 99.95 & 0.02 \\
\midrule

\multirow{6}{*}{ALBEF} &
\multirow{3}{*}{LID} &
Sep\textsubscript{uni}   & 94.99 & 23.83 & 91.80 & 35.88 \\
& & Sep\textsubscript{multi} & 74.26 & 78.75 & 79.85 & 64.51 \\
& & Co-Attack            & 93.80 & 27.98 & 91.49 & 35.58 \\
\cmidrule{2-7}
& \multirow{3}{*}{$k$-NN} &
Sep\textsubscript{uni}   & 99.75 & 1.05 & 98.54 & 5.67 \\
& & Sep\textsubscript{multi} & 54.84 & 92.46 & 57.02 & 88.27 \\
& & Co-Attack            & 99.88 & 0.50 & 98.73 & 6.84 \\
\bottomrule
\end{tabular}%
}
\end{subtable}
\begin{subtable}[t]{.48\linewidth}
\centering
\caption{Results for CLIP\textsubscript{ViT} and TCL Models}

\resizebox{\linewidth}{!}{%
\begin{tabular}{@{}lllcccc@{}}  
\toprule
\multirow{3}{*}{\textbf{Model}} &
\multirow{3}{*}{\textbf{Method}} &
\multirow{3}{*}{\textbf{Attack}} &
\multicolumn{4}{c}{\textbf{Dataset}} \\
\cmidrule(lr){4-7}
& & & \multicolumn{2}{c}{Flickr30k} & \multicolumn{2}{c}{COCO} \\
\cmidrule(lr){4-5}\cmidrule(lr){6-7}
& & & AUC & FPR95 & AUC & FPR95 \\
\midrule

\multirow{4}{*}{CLIP\textsubscript{ViT}} &
\multirow{2}{*}{LID} &
Sep\textsubscript{uni} & 99.37 & 1.51 & 99.98 & 0.20 \\
& & Co-Attack         & 96.55 & 25.63 & 99.06 & 5.08 \\
\cmidrule{2-7}
& \multirow{2}{*}{$k$-NN} &
Sep\textsubscript{uni} & 100.00 & 0.00 & 100.00 & 0.00 \\
& & Co-Attack         & 99.59  & 0.50 & 99.51  & 1.27 \\
\midrule

\multirow{6}{*}{TCL} &
\multirow{3}{*}{LID} &
Sep\textsubscript{uni}   & 90.88 & 40.93 & 89.32 & 42.52 \\
& & Sep\textsubscript{multi} & 84.72 & 56.47 & 83.95 & 58.16 \\
& & Co-Attack            & 90.76 & 37.31 & 88.25 & 43.79 \\
\cmidrule{2-7}
& \multirow{3}{*}{$k$-NN} &
Sep\textsubscript{uni}   & 96.10 & 19.60 & 98.01 & 11.24 \\
& & Sep\textsubscript{multi} & 32.89 & 95.48 & 33.89 & 95.41 \\
& & Co-Attack            & 96.59 & 15.07 & 98.05 & 11.73 \\
\bottomrule
\end{tabular}%
}
\end{subtable}

\endgroup
\end{table*}

\subsubsection{Performance of GeoDetect in Image-Text Retrieval.}
We also evaluate image-text retrieval to demonstrate that GeoDetect applies beyond classification, without labeled data. 
Due to the lack of labels, we evaluate only LID and $k$-NN distance, as they do not require class labels. As shown in \cref{nonclassification}, the performance of all models is comparable to their classification results. 
For both the COCO and Flickr30k datasets, each image is annotated with five captions. To maintain consistency, as Co-Attack requires a matching prompt to simultaneously attack both the image and the associated text, we use the first caption as the target text.

\begin{table}[t]
    \caption{GeoDetect discrimination power (AUC score) comparison between different distribution adaptive and non-adaptive attacks for Image-Retrieval Task with Flickr30k and COCO dataset in aligned VLPs (CLIP\textsubscript{CNN} and CLIP\textsubscript{ViT}), and fused VLPs (ALBEF and TCL) (Note: 'N-adaptive' refers to the Non-adaptive method.)}
    \label{adaptype2}
    \centering
    \footnotesize
    \setlength{\tabcolsep}{3pt}
    \renewcommand{\arraystretch}{1.05}

    \begin{subtable}{.49\linewidth}
      \centering
      \caption{Effect of $k$-NN adaptive Attacks}
      \begin{adjustbox}{max width=\linewidth}
      \begin{tabular}{@{}l l c c c c@{}}
        \toprule
        \multirow{2}{*}{\textbf{Model}} & \multirow{2}{*}{\textbf{Attack}} & \multicolumn{4}{c}{\textbf{Dataset}} \\ 
        \cmidrule(lr){3-6}
        & & \multicolumn{2}{c}{Flickr30k} & \multicolumn{2}{c}{COCO} \\ 
        \cmidrule(lr){3-4} \cmidrule(lr){5-6}
        & & N-adap. & Adap. & N-adap. & Adap. \\ 
        \midrule
        \multirow{2}{*}{CLIP\textsubscript{CNN}} 
        & Sep\textsubscript{uni}   & 99.99 & 67.82 & 99.97 & 74.47 \\ 
        & Co-Attack                & 99.97 & 67.32 & 99.95 & 74.28 \\ 
        \midrule
        \multirow{2}{*}{CLIP\textsubscript{ViT}} 
        & Sep\textsubscript{uni}   & 100   & 61.16 & 100   & 68.44 \\ 
        & Co-Attack                & 99.59 & 59.68 & 99.51 & 67.68\\ 
        \midrule
        \multirow{3}{*}{ALBEF} 
        & Sep\textsubscript{uni}   & 99.75 & 51.38 & 98.54 & 71.53 \\
        & Sep\textsubscript{multi} & 54.84 & 49.82 & 57.02 & 69.88 \\ 
        & Co-Attack                & 99.88 & 51.02 & 98.73 & 71.53 \\ 
        \midrule
        \multirow{3}{*}{TCL} 
        & Sep\textsubscript{uni}   & 96.10 & 51.27 & 98.01 & 74.19 \\
        & Sep\textsubscript{multi} & 32.89 & 50.57 & 33.89 & 74.04 \\ 
        & Co-Attack                & 96.59 & 51.18 & 98.05 & 74.00 \\ 
        \bottomrule
      \end{tabular}
      \end{adjustbox}
    \end{subtable}\hfill
    \begin{subtable}{.49\linewidth}
      \centering
      \caption{Effect of LID adaptive Attacks}
      \begin{adjustbox}{max width=\linewidth}
      \begin{tabular}{@{}l l c c c c@{}}
        \toprule
        \multirow{2}{*}{\textbf{Model}} & \multirow{2}{*}{\textbf{Attack}} & \multicolumn{4}{c}{\textbf{Dataset}} \\ 
        \cmidrule(lr){3-6}
        & & \multicolumn{2}{c}{Flickr30k} & \multicolumn{2}{c}{COCO} \\ 
        \cmidrule(lr){3-4} \cmidrule(lr){5-6}
        & & N-adap. & Adap. & N-adap. & Adap. \\ 
        \midrule
        \multirow{2}{*}{CLIP\textsubscript{CNN}} 
        & Sep\textsubscript{uni}   & 98.45 & 83.81 & 99.54 & 93.14 \\ 
        & Co-Attack                & 98.90 & 82.81 & 99.50 & 94.67 \\ 
        \midrule
        \multirow{2}{*}{CLIP\textsubscript{ViT}} 
        & Sep\textsubscript{uni}   & 99.37 & 31.56 & 99.98 & 83.82 \\ 
        & Co-Attack                & 96.55 & 52.24 & 99.06 & 86.96 \\ 
        \midrule
        \multirow{3}{*}{ALBEF} 
        & Sep\textsubscript{uni}   & 94.99 & 71.78 & 91.80 & 89.86 \\ 
        & Sep\textsubscript{multi} & 74.26 & 89.83 & 79.85 & 92.31\\ 
        & Co-Attack                & 93.80 & 70.12 & 91.49 & 90.11 \\ 
        \midrule
        \multirow{3}{*}{TCL} 
        & Sep\textsubscript{uni}   & 90.88 & 77.15 & 89.32 & 88.70  \\ 
        & Sep\textsubscript{multi} & 84.72 & 91.83 & 83.95 & 94.64 \\ 
        & Co-Attack                & 90.76 & 78.10 & 88.25 & 90.02 \\ 
        \bottomrule
      \end{tabular}
      \end{adjustbox}
    \end{subtable}
\end{table}

\subsection{Evaluation of Adaptive Attacks}

\label{sec:adaptive}
We evaluate GeoDetect under strong white-box adaptive attacks that explicitly incorporate the detector into the perturbation optimization. Following prior work on adaptive adversarial attacks \cite{athalye2018obfuscated,bryniarski2021evading}, we consider two types of adaptive attacks: 
(i) attacks generated from a different batch distribution than the detection batch, and 
(ii) Selective gradient descent adaptive attacks that balance misclassification and detection objectives. In the following attack setting, the batch size for CLIP\textsubscript{CNN} and CLIP\textsubscript{ViT} is set to 128, with $k = 100$ for LID and $k = 10$ for $k$-NN. 
For ALBEF and TCL, the batch size is set to 32, with $k = 20$ for LID and $k = 10$ for $k$-NN.  

\subsubsection{Different-Distribution Adaptive Attacks.}
We first evaluate adaptive attacks where the batch used for attack generation differs from the batch used for detection. In this adaptive setting, the attacker is assumed to know the VLP, the detector family, the calibration rule, and the geometric score used during optimization. However, the final evaluation uses a disjoint clean reference batch from the one used during attack generation. This prevents the attacker from exploiting the specific local neighborhood structure of a single batch and avoids misleading gradients caused by unstable $k$-nearest-neighbor estimates \cite{athalye2018obfuscated}. 
Adaptive perturbations are optimized using
\begin{equation} \mathcal{L}_{adaptive}(z_i,z'_i) = \mathcal{L}_{main}(z_i,z'_i) - \zeta \cdot \text{Metric}(z'_i, \{z_j\}_{j=1}^{n}), z_i \in B_g \neq B_d. \end{equation}

Here, $\text{Metric}(z'_i, \{z_j\}_{j=1}^{n})$ represents the LID or $k$-NN function that computes the score for AE embeddings $z'_i$ relative to the clean sample embeddings $z_i$. The batch used for attack generation, $B_g$, is different from the batch used for detection, $B_d$. We set $\zeta = 0.1$, $k=20$ for LID, and $k=10$ for $k$-NN in optimization of attacks. 
 The results presented in \cref{adaptype2} show that across aligned VLPs (CLIP) and fused VLPs (ALBEF and TCL) on Flickr30k and COCO, GeoDetect remains robust under this challenging setting. In particular, GeoDetect-LID maintains strong discrimination power with high AUC scores even when the attacker has full white-box access to the detector objective. While adaptive attacks reduce performance compared to non-adaptive settings, GeoDetect continues to reliably distinguish adversarial from clean samples across all models.

\subsubsection{Selective Gradient Adaptive Attacks.}
We also evaluate selective gradient descent adaptive attacks \cite{bryniarski2021evading}, which alternate optimization between misclassification and detection objectives to avoid local minima. These attacks provide an even stronger adaptive setting. Detailed explanation and experimental results for this attack type are reported in Appendix D.3
.

\subsection{Extended Evaluation and Ablation Study} 
Appendix B.6 
provides empirical verification of GeoDetect, including visualizations showing clear separability between clean and adversarial samples via geometric scores. Sensitivity analyses over neighborhood size, sample availability, batch size, layer choice, and multimodal layers are presented in Appendix C
, demonstrating robustness to these variations. Appendix D 
evaluates generalization to diverse attack backbones, the SGA attack~\cite{lu2023set}, and selective gradient adaptive attacks; GeoDetect remains robust. Appendix E 
reports extended evaluations on additional models (TCL, \textsc{CLIP}\textsubscript{ViT}) and a comparison with PIP~\cite{zhang2024pip} (a VQA-specific adversarial detector), showing that GeoDetect maintains its performance across models and achieves superior results to PIP. 

\section{Conclusion}
\label{conclusion}
In this paper, we propose the first task-agnostic, theoretically grounded framework for detecting AEs in VLPs. By leveraging the anisotropic structure of VLP embedding spaces, we show through theoretical analysis that adversarial perturbations push embeddings into off-manifold regions, leading to fundamental geometric differences between clean and perturbed samples. Building on this insight, we introduce GeoDetect, a lightweight, model-agnostic detection method that applies simple geometric metrics to image or joint representations. GeoDetect generalizes across multiple tasks and VLP architectures, 
and achieves strong detection performance against a range of state-of-the-art adversarial attacks. Notably, GeoDetect remains effective even under adaptive attack settings, where adversaries are aware of the detection strategy and attempt to bypass it. This robustness, combined with its independence from task-specific logits or labels, makes GeoDetect well-suited for both classification and retrieval scenarios. 
\section*{Acknowledgements}
This research was supported by The University of Melbourne’s Research Computing Services, the Petascale Campus Initiative, and the Spartan HPC facilities. This facility was established with the assistance of LIEF Grant LE170100200. Moreover, this research was supported by the ARC Centre of Excellence for Automated Decision-Making and Society (CE200100005), and partially funded by the Australian Government through the Australian Research Council.

\bibliographystyle{splncs04}
\bibliography{main}

\end{document}